# Sentence Simplification Aids Protein-Protein Interaction Extraction

Siddhartha Jonnalagadda*           Graciela Gonzalez
Department of Biomedical Informatics
Arizona State University
Phoenix, AZ 85004, USA
*sjonnal3@asu.edu

## Abstract

Accurate systems for extracting Protein-Protein Interactions (PPIs) automatically from biomedical articles can help accelerate biomedical research. Biomedical Informatics researchers are collaborating to provide meta-services and advance the state-of-art in PPI extraction. One problem often neglected by current Natural Language Processing systems is the characteristic complexity of the sentences in biomedical literature. In this paper, we report on the impact that automatic simplification of sentences has on the performance of a state-of-art PPI extraction system, showing a substantial improvement in recall (8%) when the sentence simplification method is applied, without significant impact to precision.

## 1  Introduction

The complexity of sentences characteristic to biomedical articles poses a challenge to natural language parsers, which are typically trained on large-scale corpora of non-technical text. BioInfer (Pyysalo et al., 2007) demonstrated the usefulness of even the simplest simplification steps to counter such problems; as it truncates a few dependencies related to noun phrases and also removes non-traditional characters which could potentially confuse parsing. Recently, bioSimplify (Jonnalagadda, Tari, Hakenberg, Baral, & Gonzalez, 2009) proved the utility of sentence simplification to improve the output of parsers in biomedical literature. This paper focuses on the impact of sentence simplification on a higher-level task: the extraction of associations from biomedical articles – particularly extracting protein-protein interaction (PPI). We also present improvements made to the noun phrase replacement and sentence splitting algorithms of bioSimplify and a more rational approach in using the inter-token dependencies to evaluate the complexity of the sentences. Finally, we explain the methods used to evaluate the efficacy of sentence simplification in PPI extraction and present our results.

## 2  Background

### 2.1  Protein-Protein Interaction Extraction

The study of protein-protein interactions and other molecular events is a central tenet of modern translational and genomic research. Publications centering on reports of such atomic events abound, and their manual extraction from the literature currently occupies many trained curators that deposit them in databases such as DIP, MINT, or IntAct. Manual curation, however, despite years of effort, has only made a small dent (calculated at around 7%) into the volume of publications believed to report protein-protein interactions. Automatic extraction of such facts is thus a priority for biomedical text mining researchers, although performance is still poor. The Biocreative II effort (Krallinger, Leitner, & Valencia, 2007) to compare the PPI extraction tools of 16 international teams revealed that the best system had an f-score of only 0.3. Although this performance combined extraction and gene normalization as a single task, the fact remains that there is still much to be done in this area. We attempt to increase the performance of extraction system by reducing the complexity of sentences that could be hiding PPIs.

Sentences in biomedical literature are significantly more complex than those in newspaper articles, for example, because of higher average sentence length (Jonnalagadda et al., 2009), inconsistent use of nouns and partial words (Tateisi & Tsujii, 2004), higher perplexity measures (Elhadad, 2006), greater lexical density, and increased number of relative clauses and prepositional phrases (Gemoets, 2004). The working hypothesis behind this work is that simplifying the sentences by removing these complexities could increase the performance of different extraction tools.

There are currently a few online tools for extracting PPIs, such as SPIES, Whatizit, EBIMED, RelEx, PolySearch and PIE. We have chosen PIE (Kim et al., 2008), a machine-learning based approach available as a web service and claiming 92% f-score on BioCreative II (Krallinger et al., 2007). PIE uses the parse tree information from the Collins statistical parser as its key component.

For the purpose of evaluation we use AIMed corpus (Bunescu et al., 2005) which is also created to help in comparing PPI extraction methods.

### 2.2 bioSimplify

Biomedical sentences are processed by bioSimplify in four phases:
1. Removal of spurious phrases (such as section indicators) and rule-based resolution of coordination ellipsis.
2. Replacement of gene names with single-word placeholders using BANNER (Leaman & Gonzalez, 2008), a gene named entity recognition tool.
3. Replacement of noun phrases (that don't contain genes) with single-word placeholders using LingPipe (Alias-i, 2006).
4. Syntactic splitting of a sentence into multiple sentences using inter-token dependencies.

Previously published results (Jonnlagadda, et al., 2009) testing bioSimplify on the McClosky & Charniak parser (McClosky & Charniak, 2008), one of the best open-source parsers of biomedical text, showed that pre-processing the sentences with bioSimplify improved the parser's accuracy from 76% to 79%. Similar results were observed for the Link Grammar parser, one of the best rule based parser, for which accuracy improved from 72% to 76%. This measurement was based on the Stanford dependencies calculated in the same way as in the earlier evaluation of the parsers for biomedical text (Clegg & Shepherd, 2007).

The section below describe the changes made in the last two phases of bioSimplify and the impact that subjecting the sentences to it can have on association extraction.

## 3 Methods

### 3.1 Replacement of noun phrases

bioSimplify replaces the noun phrases without gene names with meaningful single-word place holders. Consider the sentence "Mutations in CBP have recently been identified in RTS patients." The original bioSimplify simplified it to "Mutations in GENE0 have recently been identified in REPNP0." Replacement of gene names with place holders like GENE0 does not generally lead to loss of context for the task of PPI extraction, as we maintain a list of place holders cross-referenced to the corresponding original gene names. However, replacing noun phrases with place holders like REPNP0 can cause loss of context because of skipping the words that indicate association. Hence, we replace the noun phrases with the head noun of the phrase. For example, the "RTS patients" is replaced by "patients" instead of the abbreviation REPNP0.

### 3.2 Syntactic Simplification

The problem of simplifying long sentences in common English text has been studied before, notably by (Chandrasekar & Srinivas, 1997), (Carroll, et al., 1998) in their Practical Simplification of English text project or PSET and (Siddharthan, 2006). The goal of syntactic simplification is to reduce the grammatical complexity of a text while retaining the relevant information content and meaning to enable better processing by parsers, and better readability for humans. This work and that of (Chandrasekar & Srinivas, 1997) cater to better processing by parsers, while (Carroll et al, 1998) and (Siddharthan, 2006) cater to better readability for humans. Our naive approach to syntactic simplification (Jonnalagadda et al., 2009) was a first step, but it did not have a method to determine whether a simplified sentence is grammatically correct or not. In this paper we first address the problem of accurately determining the grammatical correctness of a sentence and then describe our rule-based approach for splitting sentences inspired by (Siddharthan, 2006).

### 3.3 Grammatical correctness of a sentence

For the sentence "*He is an amazing.*", the Stanford parser gives a complete constituent tree as if it were a grammatically correct sentence, while the Link Grammar parser (Sleator, 1998) correctly gives an incomplete linkage leaving behind "an", thus saying that "*He is amazing.*" is the closest grammatically correct sentence. However, Link

Grammar gives too many false negatives. In fact, Link Grammar doesn't find a complete linkage for more than 33% of the 1100 sentences in BioInfer corpus, all of which are grammatically correct (Jonnalagadda et al., 2009).

The statistical parsers generally give more false-positives than dependency parsers as they try to make a 'best guess' at a badly-constructed sentence, given that robustness in the face of poor grammar is more desirable. Most of the evaluation of the performance of parsers (for example, (Clegg & Shepherd, 2007)) is done using a corpus of grammatically correct English sentences. Those results are not relevant in our context where the primary task is to determine the correctness of a sentence. (Foster, 2007) uses an automatically generated Treebank of grammatically incorrect sentences, but rightly notes that the correct combination of grammatically correct and incorrect training data for a statistical parser has yet to be created. The statistical parsers trained using only grammatically correct sentences will not be able to catch errors not found in the training data.

(Siddharthan, 2006) comments that it is difficult to automatically determine the grammatical correctness of a sentence; so he evaluates them manually. However, it will be very useful for the task of simplification to have an efficient method to automatically compare the syntactical soundness of the simplified parts to the original part before committing the simplification. We propose using the combination of the number of null links and disjunct cost from the cost vector returned by the Link Grammar (version: 4.5.3) to get an estimate of the grammatical correctness.

The vector "UNUSED" and "DIS" in the cost vector indicate the number of null links and disjunct cost, respectively. Figure 1 is the output for the sentence "*This is an amazing*". Here "an" from the original sentence is removed to suggest the closest grammatically correct sentence.

Figure 2 is the output for the sentence "*This is an dangerously.*" We see that the null count for both the sentences is 1, but the suggestion for the second sentence – "*This is dangerously.*" is less grammatically correct than the suggestion for the first sentence – "*This is amazing.*". This difference in grammatical correctness is reflected by the disjunct cost or DIS vector; the DIS vector for the second sentence is 2 while that for the first sentence is 0. Disjunct cost, also called Connector cost, represents the level of inappropriateness (caused by using less frequent rules) in the linkage.

Fig 1: Link Grammar's output

Fig 2: Link Grammar's another output

Every sentence can be uniquely associated with the 2-tuple of null count and disjunct cost. It is reasonable to assume that a null count (which represents unwanted words) needs more attention than the disjunct cost (which represents less likely linkages). Since null counts and disjunct costs are typically less than 10 (i.e, one-digit numbers), for the purpose of easy comparison and for capturing the 2-tuples in one dimension, we define a new cost vector GRAM which is equal to 10*UNUSED + DIS. It is an easy proof that GRAM value is equivalent to the 2-tuple of null count and disjunct cost, under the assumption that the disjunct cost of the corresponding collection of sentences is not more than 10.

Any syntactic simplification will be approved only if the resulting sentences are collectively at least as grammatically correct as the original sentence alone, i.e, the sum of GRAM values of the parts should be less than or equal to the GRAM value of the original sentence. For example, the GRAM value of the sentence – "*These effects were associated with significantly lower blood pressure, though within the normal range, in captopril-treated versus control animals.*" is 20 (UNUSED = 2, DIS = 0; skipped "though" and "versus"). One suggestion from bioSimplify for splitting into the two sentences is: "*These effects were associated with significantly lower blood pressure.*" and "*Within the normal range, in captopril-treated versus control animals.*", whose respective GRAM values are 0 (UNUSED =0, DIS =0) and 22 (UNUSED =2, DIS =2). Since the sum of GRAM values of the parts (22) is more than the GRAM value of the original sentence (20), this suggestion is rejected. This approximates how a human would discern, because the second sentence was not grammatically correct. The second suggestion from bioSimplify was to split it into the two sentences – "*These effects were associated with significantly lower blood pressure in captopril-treated versus control animals.*" and "*Significantly lower blood pressure is though within the normal range.*", whose respective GRAM values are 10 (UNUSED =1, DIS =0; skipped "versus") and 10 (UNUSED =1, DIS =0; skipped "though"). Since the sum of GRAM values of the parts (20) is same as the GRAM value of the original sentence (20), this suggestion is accepted. In fact, if the second sentence was instead "*This blood pressure is still within the normal range*," the GRAM value of the constituents would be lesser than that of the original sentence. This would also be possible in the future with more advanced implementations for resolving relative and appositive clauses in the syntactic simplification.

### 3.4 Overview of Rules for simplification

We implemented the rules for prefix subordination, infix subordination and if-then coordination (details in Siddharthan, 2003). These rules were also adapted recently by SimText (Ong, et al., 2008), a text simplification system for improving the readability of medical literature, but without a mechanism to judge the grammatical correctness. We add the notion of the GRAM value (as described in the prior section) as a way to automatically judge the syntactical soundness of the simplified parts as compared to the original sentence. There are seven rules in total – three for conjunction and two each for relative clauses and apposition.

The referring expression (Siddharthan, 2003) for the relative clause is determined using the "MX" link from the link grammar output and that for appositive clauses is determined using the "R" link from the link grammar output. In addition, all the describing phrases that occasionally occur at the beginning of the sentence like the underlined phrases in: "*These results suggest that affixin is involved in reorganization of subsarcolemmal cytoskeletal actin by activation of Rac1 through alpha and betaPIXs in skeletal muscle*." and "*As reported previously, alphaPIX was specifically co-immunoprecipitated by anti-affixin and anti-betaPIX antibodies*." are removed.

## 4 Evaluation

The BCMS platform (Leitner et al., 2008) provides meta-services for information extraction in molecular biology. There are 12 research labs providing BCMS, but currently (as the outcome of BioCreative II) the publicly available servers only give information on whether the abstract with a given PubMed ID contains at least one PPI or not. We are studying whether simplification of a sentence helps PPI extraction systems, so it was more appropriate to use a tool which operates on single sentences. For this reason, and the fact that it uses parse tree information, PIE (Kim et al., 2008) was selected as an ideal tool for evaluating bioSimplify. PIE is available as an online web service that can test any sentence(s) for presence of PPIs, not just a PubMed abstract. More information about the usage of PIE is available at http://bi.snu.ac.kr/pie. PIE returns positive or negative for each sentence depending on whether or not it detects a PPI in it.

Table 1 shows some examples where bioSimplify simplified a sentence and helped correct the output of PIE. The sentences in which PIE reported a PPI are marked in **bold**.

Table 1: Examples of impact of bioSimplify in improving association extraction

| Original Sentence | Simplified Sentence | Comment |
|---|---|---|
| Unlike human IL - 6 ( which uses many hydrophilic residues ), the viral cytokine largely uses hydrophobic amino acids to contact gp130, which enhances the complementarity of the viral IL - 6 - gp130 binding interfaces. | Unlike human IL - 6, the viral cytokine largely uses acids to contact gp130.<br>**gp130 enhances the complementarity of the viral IL - 6 - gp130 binding interfaces.** | False negative converted to True positive. bioSimplify has split the sentences into two parts (in addition to preprocessing and replacing noun phrases) and made it easy for PIE to identify the PPI. |
| LEC also induced calcium mobilization, but marginal chemotaxis via CCR5. | **LEC also induced calcium mobilization, but chemotaxis via CCR5.** | Replacing noun phrase with single word helps to concentrate on PPI indicating words and structure |
| The sequences that confer on FGF - 7 its specific binding to KGFR have not been identified. | **The sequences confer on FGF - 7 its specific binding to KGFR .**<br>The sequences have not been identified. | Splitting sentences again uncovers the PPI indicated in part of the sentence. |
| It has been shown that LIGHT triggers apoptosis of various tumor cells including HT29 cells that express both lymphotoxin beta receptor ( LTbe- | LIGHT triggers apoptosis of cells including cells.<br>**Cells express both lymphotoxin beta recep-** | Noun phrase replacement, preprocessing and sentence splitting together separate the PPI containing part of the sentence for easy identi- |
| taR ) and HVEM / TR2 receptors. | tor and HVEM / TR2 receptors. | fication |
| Thus, Phe93 and Phe205 are important binding determinants for both EPO and EMP1, even though these ligands share no sequence or structural homology, suggesting that these residues may represent a minimum epitope on the EPOR for productive ligand binding. | **Thus, Phe93 and Phe205 are determinants for both EPO and EMP1.** These ligands share no sequence or structural homology, suggesting that these residues may represent a minimum epitope on the EPOR for binding. | Splitting sentences again uncovers the PPI indicating part of the sentence. |
| We have crystallized a complex between human FGF1 and a two - domain extracellular fragment of human FGFR2. | **We have crystallized a complex between human FGF1 and a two - domain fragment of human FGFR2.** | Even a little amount of simplification sometimes highly influences the PPI extraction. |
| **The structural arrangement in the active site is consistent with a mostly associative mechanism of phosphoryl transfer and provides an explanation for the activation of Ras by glycine - 12 and glutamine - 61 mutations.** | The structural arrangement in the active site is consistent with a mostly associative mechanism of transfer. Arrangement provides an explanation for the activation of Ras by glycine - 12 and glutamine - mutations. | False positive converted to true negative. bioSimplify transformed the sentence and helped PIE in deciding that there is actually no PPI in the sentence, though there seems to be one because of the complexity of the sentence. |

The AIMed corpus contains annotation for 197 abstracts which were identified by the Database of Interacting Proteins (DIP) to have PPIs and 29 more which don't have PPIs. In each sentence, all the proteins and all the pair-wise interactions among them are annotated. AIMed is publicly available at ftp://ftp.cs.utexas.edu/pub/mooney/biodata/interactions.tar.gz. Figure 3 explains the evaluation steps.

Considering a sentence as a positive if PIE detects a PPI in either the original sentence or in its simplified version should, in theory, lead to a system with higher recall without negatively affecting precision. For explanation, let us assume this set-up. bioSimplify transforms each sentence into at least one sentence that could be different from the original. Suppose it transforms a sentence A into A1,…,An for n>0. There are four possible outcomes: a) <u>A is falsely assigned positive by PIE</u>: Irrespective of whether A1,…,An are falsely assigned positive, there is no change in the precision and recall. b) <u>A is correctly assigned negative</u>: It is highly unlikely that the system which performed well with A, commits a mistake on the simpler parts – A1,…,An; so the precision is less likely to decrease and the recall would be the same irrespective of the performance of the system on A1,…,An. c) <u>A is correctly assigned positive</u>: Irrespective of the performance of the system on A1,…,An, there is no change in the precision and recall. d) <u>A is falsely assigned negative</u>: Though the system failed to identify PPI in A, the simplified segments A1,…,An could allow a PPI extraction system to identify a PPI. Thus, addressing the false negatives (case d) without increasing false positives (case b) would increase both the recall and precision of the system. Cases a) and c) don't affect the system's performance. Case b presents a slight likelihood for a decrease in precision (where the simplified sentence triggers a false positive), but because this (in theory) happens sparsely -we detected this in 2% of the cases while processing the AIMed corpus- and there is a high likelihood of increase in recall and precision by resolving false negatives (case d), overall the system would have a higher recall with almost the same precision when processing simplified sentences. We refer to this set-up as **OR combination**. An alternate set-up where a sentence is considered positive only when PIE identifies a PPI in both the original sentence and the corresponding simplified sentence would be referred to as **AND combination.**

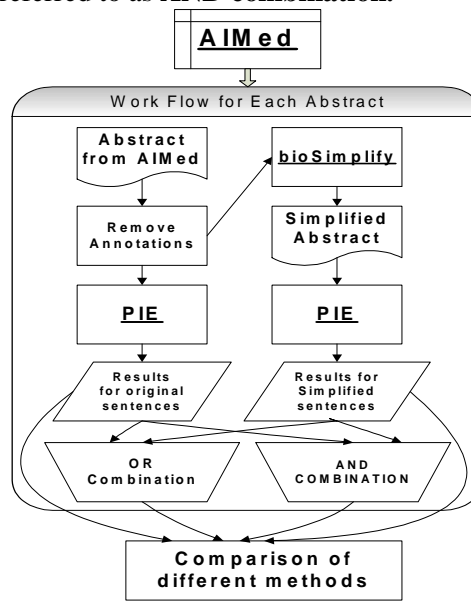

Fig 4: Evaluating impact of sentence simplification.

We used PIE to test for the presence of PPIs in 942 sentences before and after simplification. PIE reports whether a sentence with a potential PPI has a high probability of having a PPI or just a mod-erate probability. Any sentence in which PIE identifies a PPI with at least a moderate probability is considered as positive. These sentences are from 76 PubMed abstracts in AIMed with ids between 8816798 and 11470772 (this selection was based on DIP), and 14 PubMed abstracts chosen with PubMed ids between 11780382 and 11790884 for negative examples of interaction. Overall, out of the 942 sentences in these abstracts, 270 contain PPI(s). Each abstract was processed as illustrated in Figure 3. The aggregate results of PIE on all the sentences and their simplified counterparts are presented in Table 2.

As was postulated, the highest f-score is observed for OR combination, with recall increasing by 8% with almost the same precision and the f-score is 3% more than that for the original 942 sentences. These results show that sentence simplification can have a substantial positive impact on the extraction of PPIs.

Table 2: Results of PIE on selection from AIMed

| Category | Recall | Precision | f-score |
| --- | --- | --- | --- |
| Before simplification | 53% | 49% | 51% |
| After simplification | 55% | 52% | 53% |
| AND combination | 47% | 53% | 50% |
| **OR combination** | **61%** | **48%** | **54%** |

## 5 Conclusion

The results of evaluation and error analysis allow us to conclude that bioSimplify, although still needing improvements, leads to improved PPI extraction results using PIE, which already uses syntactic information from parse trees. The results indicate that a system for sentence simplification used as a preprocessing step for natural-language processing PPI extraction systems could improve the PPI extraction process and other association extraction processes.

## Acknowledgments

We thank Science Foundation Arizona (award CAA 0277-08 Gonzalez) for partly supporting this research.